\documentclass{article}

\usepackage[preprint]{neurips_2026}

\usepackage[utf8]{inputenc}
\usepackage[T1]{fontenc}
\ifdefined\XeTeXversion
  \catcode"2019=\active
  \begingroup
    \lccode`\~="2019
    \lowercase{\endgroup
      \def~{\textquoteright}}
\else
  \DeclareUnicodeCharacter{2019}{\textquoteright}
\fi
\usepackage{amsmath}
\usepackage{amssymb}
\usepackage{booktabs}
\usepackage{graphicx}
\usepackage{hyperref}
\usepackage{microtype}
\usepackage{xcolor}
\usepackage{mathptmx}

\hypersetup{
  colorlinks=true,
  linkcolor=blue,
  citecolor=blue,
  urlcolor=blue
}

\definecolor{MistralOrange}{HTML}{FF7000}
\definecolor{TetherPurple}{HTML}{542F91}
\definecolor{PVFTeal}{HTML}{006D77}
\newcommand{\lecritique}{\textcolor{MistralOrange}{\textsc{Le Critique}}}

\newcommand{\tether}{\textcolor{TetherPurple}{\textsc{Tether}}}
\newcommand{\pvf}{\textcolor{PVFTeal}{\textsc{Pvf}}}
\newcommand{\E}{\mathbb{E}}
\newcommand{\Var}{\operatorname{Var}}
\newcommand{\clip}{\operatorname{clip}}
\newcommand{\loo}{\textsc{Loo}}

\title{\lecritique: Privileged Value Functions\\ for\\ LLM Reinforcement Learning}

\author{%
  Siddarth Venkatraman\textsuperscript{1,2,3}
  \qquad
  Matthieu Dinot\textsuperscript{1}
  \qquad
  Laurence Aitchison\textsuperscript{1}\\[0.6em]
  {\normalfont\small
    \textsuperscript{1}Mistral AI
    \quad
    \textsuperscript{2}Mila -- Quebec AI Institute
    \quad
    \textsuperscript{3}Universit\'e de Montr\'eal}
}

\begin{document}

\maketitle

\begin{abstract}
  Reinforcement learning algorithms for Large Language Models (LLMs) are largely distinguished by their variance reduction strategy. Group-relative methods like GRPO reduce gradient variance by sampling multiple rollouts per prompt, but provide only sequence-level credit. Training is also blocked by straggler rollouts, reducing throughput and increasing off-policyness. Learned value functions theoretically address both problems, providing token-level advantages without requiring large groups. However, additional infrastructure engineering challenges combined with the practical success of critic-free methods have made it difficult to justify their inclusion in RL pipelines. We propose two complementary strategies to improve the performance of value function RL: 1) \textit{Privileged Value Functions} (\pvf) which provide an elegant mechanism to inject additional task-relevant token-level signal without biasing the policy objective; 2) \tether, a baseline that adaptively interpolates between group-relative and value baselines depending on the value function accuracy. Across several reasoning tasks, both strategies consistently improve over the standard value function baseline, and are competitive with or outperform mean-baseline GRPO. The asynchronous value function RL infra used for our experiments can be found \href{https://github.com/HyperPotatoNeo/prime-values}{here}.

\end{abstract}

\section{Introduction}
\label{sec:introduction}

Value functions are foundational objects in reinforcement learning \citep{sutton2018reinforcement}. Their purpose is to amortize return prediction: given a state, or a state-action pair, they estimate the expected future return without expensive and noisy Monte-Carlo sampling. These estimates support value-based control, as in
DQN \citep{mnih2013playing,mnih2015dqn,vanhasselt2016deep}, and serve the role of critics in actor-critic methods \citep{mnih2016a3c, lillicrap2016ddpg, schulman2017ppo, haarnoja2018sac}. The critic (term we use interchangeably with value function) serves both as a variance-reducing baseline for policy gradient estimation and as a source of bootstrap targets for temporal credit assignment. Value functions also facilitate efficient model-based planning as demonstrated in some of deep RL's historic successes, most notably AlphaGo and AlphaZero, which both used a value network alongside policy
networks to guide Monte Carlo tree search \citep{silver2016alphago, silver2017mastering,silver2017alphagozero}.

Value functions were also a standard component of language model RL. PPO for RLHF paired the LLM policy with a trained critic, typically implemented as a copy of the LLM with a scalar value head \citep{stiennon2020summarize,ouyang2022instructgpt,bai2022helpfulharmless,touvron2023llama2}. LLM RL has however shifted toward critic-free methods like GRPO that estimate advantages from groups of sampled responses \citep{shao2024deepseekmath, deepseekai2025r1, ahmadian2024rloo}. These methods avoid the infrastructural and algorithmic burden of maintaining a separate value model and are often more performant when the critic is poorly fitted. The empirical success of group-relative methods comes at the theoretical cost of discarding temporally fine-grained credit assignment from value functions. Large groups also exacerbate straggler effects -- training must wait for the slowest response in each group. In asynchronous RL this worsens off-policyness \citep{khan2026fastersynchronousonpolicyrl, hou2026singleasync}.
 This tension motivates our search for methods that better exploit the gains offered by value functions in LLM RL, such that the added infra costs over critic-free methods are justified.

\begin{figure}[t]
  \centering
  \includegraphics[width=\linewidth]{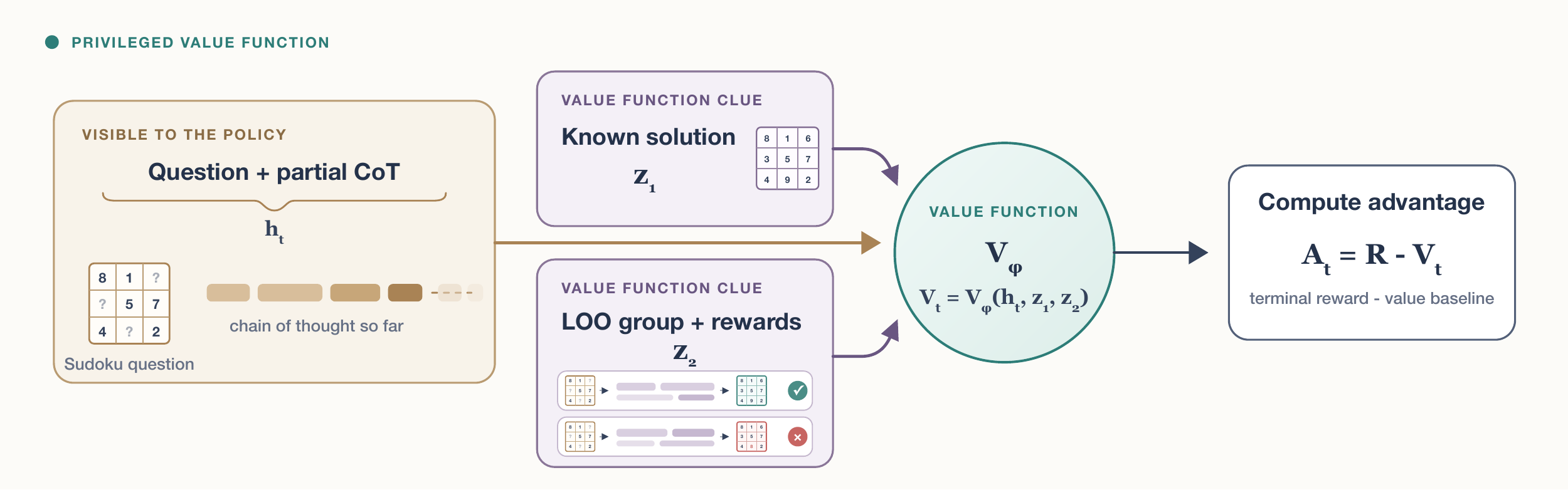}
  \caption{\textbf{Privileged value functions.} Let $h_t$ be the policy state at time $t$, here represented as a Sudoku question and an intermediate chain of thought. To improve value estimation, a Privileged Value Function (\pvf) additionally uses context clues unavailable to the policy, such as a known
  reference solution $z_1$ or leave-one-out group trajectories with rewards $z_2$. In this illustrative figure the value
  function conditions on both, producing the baseline
  $V_t=V_\phi(h_t,z_1,z_2)$ and unbiased policy advantage $A_t=R-V_t$.}
  \label{fig:privileged-value-overview}
\end{figure}

Our proposed strategy is to exploit an underexplored feature of value functions: their ability to leverage information unavailable to the policy (including group info). We introduce two methods to construct stronger value-based advantages, and demonstrate improved RL training across several reasoning-heavy tasks.

\begin{center}
\begingroup
\setlength{\fboxsep}{7pt}
\fcolorbox{black!18}{black!2}{%
\begin{minipage}{0.925\linewidth}
  \textcolor{MistralOrange}{\textbf{\textsc{Our contributions}}}
  \vspace{0.45em}

  {\setlength{\fboxsep}{5pt}%
  \colorbox{PVFTeal!8}{%
  \parbox{\dimexpr\linewidth-2\fboxsep\relax}{%
    \textcolor{PVFTeal}{\textbf{1. Privileged value functions}}
    \hfill {\small\textcolor{gray}{Section~\ref{sec:privileged}}}

    \vspace{0.2em}
    Value functions can condition on more than the current LLM state. We
    introduce \textit{Privileged Value Functions} (\pvf s), which use
    additional information hidden from the policy to improve value estimation
    and, in turn, policy optimization.
  }}}

  \vspace{0.5em}

  {\setlength{\fboxsep}{5pt}%
  \colorbox{TetherPurple!7}{%
  \parbox{\dimexpr\linewidth-2\fboxsep\relax}{%
    \textcolor{TetherPurple}{\textbf{2. Adaptive group--value baselines}}
    \hfill {\small\textcolor{gray}{Section~\ref{sec:tether}}}

    \vspace{0.2em}
    Value baselines can fail when the critic is poorly fitted, while the group
    baseline remains reliable but lacks token-level credit. We introduce
    \tether{}, which adaptively combines the group and value baselines and
    smoothly interpolates between their complementary strengths.
  }}}
\end{minipage}}
\endgroup
\end{center}

\suppressfloats[t]
\begin{figure}[t]
  \centering
  \includegraphics[width=\linewidth]{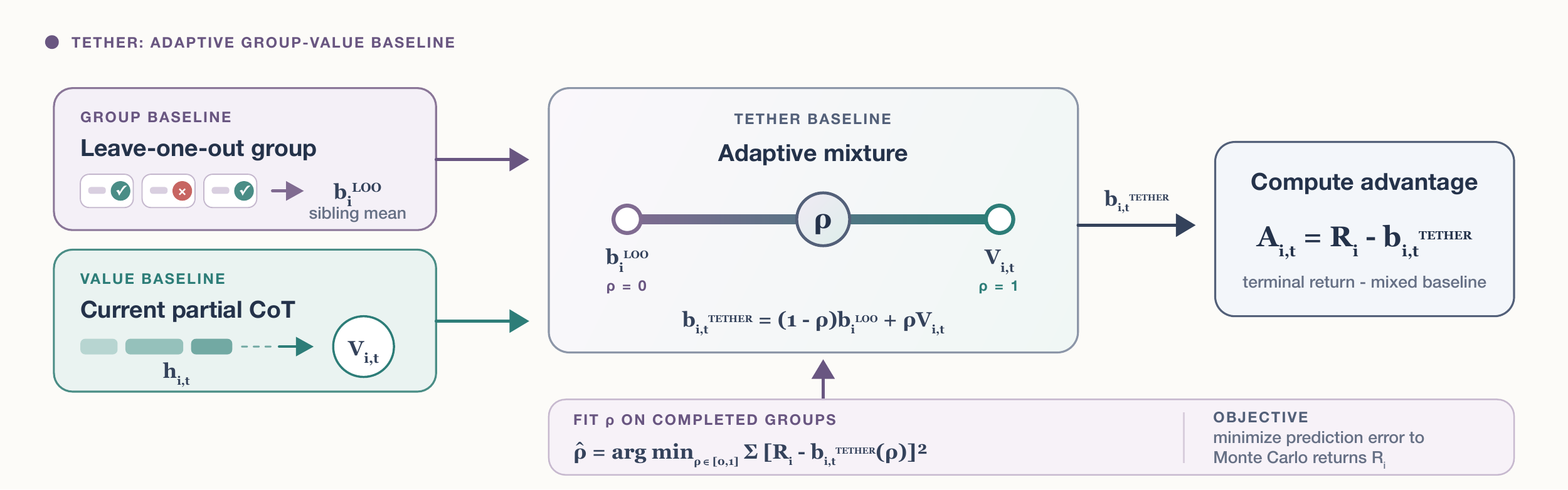}
  \caption{\textbf{\tether{}: adaptively combining group and value baselines.}
  Our other contribution is a baseline that combines group Leave-One-Out mean and token level values through simple linear combination using an adaptive mixture ratio. The coefficient
  $\rho$ is fit to minimize Monte Carlo return prediction error. $\rho=0$ recovers the group
  baseline, $\rho=1$ recovers the value function baseline, and intermediate values smoothly interpolate between the two endpoints.}
  \label{fig:tether-overview}
\end{figure}

\section{RL preliminaries}
\label{sec:background}

This section briefly introduces the necessary preliminaries for LLM RL and reviews value function concepts relevant to our discussion.

\subsection{LLM policy gradients}

LLM RL algorithms all share the same basic policy gradient loop. At each step, the policy samples one or more responses for a batch of tasks, and each response token is assigned an advantage $\widehat{A}_{i,t}$ that determines whether its probability should increase or decrease. Algorithms differ primarily in how these advantages are estimated and how the resulting policy update is stabilized.

Let a rollout batch contain $B$ sampled responses
$\tau_i=(y_{i,1},\ldots,y_{i,T_i})$ to prompts $x_i$, with
$h_{i,t}=(x_i,y_{i,<t})$ being the LLM context for response $i$ at timestep $t$. The generic token-normalized policy-gradient estimator
can be written as
\begin{equation}
  \widehat{\nabla_\theta J}
  = \frac{1}{\sum_{i=1}^{B} T_i}
    \sum_{i=1}^{B}\sum_{t=1}^{T_i}
    w_{i,t}\,\widehat A_{i,t}\,
    \nabla_\theta\log\pi_\theta(y_{i,t}\mid h_{i,t}),
  \label{eq:policy-gradient}
\end{equation}
where $\widehat A_{i,t}$ is the advantage assigned to token $y_{i,t}$. In this form, advantages do not propagate gradients back to the policy and are fixed multipliers. The stability weights $w_{i,t}$ are effective multipliers induced by the specific policy-optimization rule,
including standardization, importance ratios, clipping, or masking \citep{minimax2025m1, roux2025tapered}. There can be additional terms contributing to the gradient, like KL regularization, which we omit here for simplicity.

Value functions do
not change the policy gradient structure and determine only how the advantages $\widehat{A}_{i,t}$ are constructed. PPO estimates token-level advantages using
a learned value function, typically through Generalized Advantage Estimation (GAE)
\citep{schulman2015gae,schulman2017ppo}. GRPO instead estimates advantages by
comparing rewards across multiple responses sampled for the same prompt
\citep{shao2024deepseekmath}. The following subsections describe these group-relative and value-based advantage estimators.

\subsection{Group-relative baselines}

Group-relative methods estimate prompt difficulty from several responses to the
same prompt. For a response $\tau_i$ with scalar return $R_i$, advantages are computed using the prompt-specific baseline $b_i$:
\begin{equation}
  A_i=R_i-b_i.
  \label{eq:group-advantage}
\end{equation}

GRPO \citep{shao2024deepseekmath} estimates $b_i$ with the mean return of a group of $K$ responses, thus centering the advantages. The original formulation also normalizes the advantages, a stability detail we ignore for now:
\begin{equation}
  \bar R=\frac{1}{K}\sum_{j=1}^{K}R_j,
  \qquad A_i^{\mathrm{GRPO}}=R_i-\bar R.
  \label{eq:group-mean}
\end{equation}
Responses
with return above the mean are pushed up and those below pushed down. The advantages are sequence-level, so $A_i$ is repeated across every token in sequence $\tau_i$. A sequence's own return $R_i$ is used to compute $\bar R$, technically leading to a biased estimator. RLOO \citep{ahmadian2024rloo} removes this dependence by
forming a leave-one-out baseline from the $K-1$ sibling returns:
\begin{equation}
  b_i^{\mathrm{LOO}}=\frac{1}{K-1}\sum_{j\ne i}R_j,
  \qquad A_i^{\mathrm{LOO}}=R_i-b_i^{\mathrm{LOO}}.
  \label{eq:loo}
\end{equation}
A baseline preserves an unbiased policy gradient when its contribution to the expected score is zero. For a baseline $b(h_{i,t},z_{i,t})$, a sufficient (not necessary) condition is that the auxiliary information $z_{i,t}$ is conditionally independent of the current token given its history:
\begin{equation}
  \E\!\left[
    b(h_{i,t},z_{i,t})
    \nabla_\theta\log\pi_\theta(y_{i,t}\mid h_{i,t})
    \mid h_{i,t}
  \right]=0,
  \qquad
  z_{i,t}\;\perp\!\!\!\perp\;y_{i,t}\mid h_{i,t}.
  \label{eq:baseline-identity}
\end{equation}
The conditioning $z_{i,t}$ cannot contain future tokens, realized rewards, or later environment or verifier feedback generated by trajectory $i$, since these can depend on $y_{i,t}$ given its history. This condition will be revisited when we discuss \pvf s in Section~\ref{sec:privileged}.
  Unbiasedness is typically desirable but the GRPO baseline bias is luckily not too problematic since the GRPO and RLOO advantages differ only by constant rescaling:
\begin{equation}
  A_i^{\mathrm{GRPO}}=\frac{K-1}{K}A_i^{\mathrm{LOO}}.
  \label{eq:grpo-loo-rescaling}
\end{equation}

\subsection{Value functions and value baselines}
\label{sec:value-summary}

Value functions predict the expected return of sequences sampled by the policy $\pi$ following the token history:
\begin{equation}
  V^\pi(h_{i,t})
  =\E_{\tau\sim\pi}\!\left[R_i\mid h_{i,t}\right].
\end{equation}
The value of the task prompt $V^\pi(x_i)$ equals the expected LOO baseline (Equation~\ref{eq:loo}). At later prefixes, it tracks how the
expected reward changes as tokens are generated. A value function can
therefore provide token-level credit, and also operate without rollout groups $K=1$.

For simplicity, we focus on the setting where each sequence receives a single terminal reward $R_i$. We also assume undiscounted returns
($\gamma=1$). The simplest Monte-Carlo (MC) setting trains the value function conditioned on every prefix to predict the corresponding sequence reward. Unbiased advantages can then be computed using this value function as the token-level baseline.
\begin{equation}
  \begin{aligned}
    \mathcal{L}_V(\phi)
    &=\sum_{i,t}\left(V_\phi(h_{i,t})-R_i\right)^2,
    \qquad
    \widehat A_{i,t}=R_i-V_\phi(h_{i,t}).
  \end{aligned}
  \label{eq:mc-value}
\end{equation}
This Monte Carlo target is an unbiased sample of the conditional expectation
defining $V^\pi$, although it may suffer from high variance. An imperfectly trained value baseline still yields an unbiased policy gradient but may yield noisier gradients.

Generalizing this, Equation~\ref{eq:mc-value} is the $\lambda=1$ endpoint of the broader
temporal-difference (TD) and generalized advantage estimation (GAE) framework \citep{schulman2015gae}.
For general intermediate rewards $r_{i,t}$ (which is $0$ in our case except at the terminal step) and discount $\gamma$, the one-step
TD residual is
\begin{equation}
  \delta_{i,t}
  =r_{i,t}+\gamma V(h_{i,t+1})-V(h_{i,t}).
\end{equation}
It measures the difference between the current prediction and the value function that
observes one extra step before bootstrapping from that step. The
$\lambda$-return mixes such targets over longer horizons:
\begin{equation}
  R_{i,t}^{\lambda}
  =V(h_{i,t})
   +\sum_{l\geq0}(\gamma\lambda)^l\delta_{i,t+l}.
  \label{eq:lambda-return}
\end{equation}
$\lambda=0$ is the one-step bootstrap target
$r_{i,t}+\gamma V(h_{i,t+1})$. Increasing $\lambda$ incorporates more observed
rewards before bootstrapping, and at $\lambda=1$ the residuals telescope to the unbiased MC return. In general, increasing $\lambda$ reduces bias at the cost of variance. GAE constructs the corresponding policy advantage from these residuals:
\begin{equation}
  \widehat A_{i,t}^{\lambda}
  =\sum_{l\geq0}(\gamma\lambda)^l\delta_{i,t+l}
  =R_{i,t}^{\lambda}-V(h_{i,t}).
  \label{eq:gae}
\end{equation}
Lower $\lambda$ relies more strongly on the
critic, and with LLM RL should generally be avoided since this introduces training bias. The critic training target and policy advantage may
use separate coefficients, denoted $\lambda_{\mathrm{target}}$ and
$\lambda_{\mathrm{GAE}}$ \citep{yue2025vapo}. In our experiments, we always set $\lambda_{\mathrm{target}}=\lambda_{\mathrm{GAE}}=1.0$; i.e., advantages yield unbiased policy gradients, and value models are trained with MC targets.

\section{Privileged value functions}
\label{sec:privileged}

It is common to have privileged information during RL which the policy cannot directly condition on, but would provide useful training signal if it could somehow be injected during training. This information could be oracle answers, verifier rubrics, latent environment states, or even the other leave-one-out responses and corresponding rewards from the task group. We propose a general mechanism for incorporating such privileged information into policy training without altering the policy-gradient objective.

The key idea is to route this information through the value function. A \emph{privileged value function} (\pvf) conditions the critic on both the standard policy token history and additional training-time context. Critic access to appropriate privileged information helps it predict the return, which then improves value estimation thereby reducing policy gradient variance. In this section, we formalize the method, describe how to practically instantiate it across some standard LLM task types, and in Section~\ref{sec:pvf-experiments} empirically demonstrate that training with \pvf s consistently improves policy performance.

\subsection{Privileged information as a critic input}

Conditioning on privileged information should improve the critic's predictive performance without changing its functional role. Let
$z_{i,t}$ denote privileged context available when scoring token $t$. We
define
\begin{equation}
  V^\pi(h_{i,t},z_{i,t})
  := \E_{\tau_i\sim\pi}\!\left[
    R_i \mid h_{i,t},z_{i,t}
  \right],
  \qquad
  \widehat A^{\mathrm{PVF}}_{i,t}
  = R_i-V_\phi(h_{i,t},z_{i,t}),
  \label{eq:privileged-value}
\end{equation}
where $V_\phi$ is the \pvf{} approximating $V^{\pi}$. The policy gradient estimator using $\widehat A^{\mathrm{PVF}}_{i,t}$ remains unbiased when the privileged context satisfies the baseline admissibility condition in Equation~\ref{eq:baseline-identity}.
For example, a fixed
reference answer is allowed privileged information, whereas future tokens, realized rewards, or subsequent environment or verifier feedback generated by the current response are not. More informative conditioning in general produces a better baseline through variance reduction since
\begin{equation}
  \E\!\left[
    \left(R_i-\E[R_i\mid h_{i,t},z_{i,t}]\right)^2
  \right]
  \leq
  \E\!\left[
    \left(R_i-\E[R_i\mid h_{i,t}]\right)^2
  \right].
  \label{eq:privileged-value-mse}
\end{equation}
It is important to note the guarantee in Equation~\ref{eq:privileged-value-mse} concerns an optimal predictor, while the model $V_{\phi}$ may benefit only
if it can learn to use the additional information. For example, adding too much extra context to the \pvf{} may degrade performance even if the information should be helpful in theory. On the other hand, additional context can practically help even when already recoverable in principle from the policy context (no information gain), if it's in a form easier for the value model to represent.

Finally, we note that our work is not the first to observe that value functions can be conditioned on privileged information. This idea has been investigated before in the context of asymmetric actor--critic algorithms applied to simulated robotics control, which we credit in Appendix~\ref{app:related-work}. Our contribution is applying the idea to value function learning for LLM policy gradients.


\subsection{Available forms of privileged information}

The specific form of available privileged information is very task dependent and should be thought about carefully for every environment. An important high level question to consider is if the information could help predict the future return of a partial trajectory, simplifying the value model's job. We consider some important cases:
\paragraph{Reference solutions.}
Reference solutions provide particularly useful privileged context for a value function, since they reduce the prediction problem to assessing whether the current partial trajectory is progressing toward a known correct end state. Examples include oracle answers, proof sketches for mathematical problems, and gold patches for code-repair tasks.
\paragraph{Leave-One-Out group.}
Other group responses provide a generically available source of privileged information, including for tasks without reference solutions. Equation~\ref{eq:baseline-identity} excludes future tokens, rewards, and feedback generated by trajectory $i$, but freely permits conditioning on the other $K-1$ independently sampled responses and even their rewards. This effectively provides the critic with an aggregation prompt simplifying value prediction to an in-context learning task, a strong inductive bias in prior work \citep{venkatraman2026recursiveselfaggregationunlocksdeep,singh2026v_1}. We also observe from this perspective, the LOO baseline in Equation~\ref{eq:loo} can itself be viewed as a training-free \pvf{} conditioned only on the group rewards:
\begin{equation}
  V_{\phi}\!\left(h_{i,t},\{R_j\}_{j\neq i}\right)
  =
  \frac{1}{K-1}\sum_{j\neq i}R_j.
\end{equation}
This insight will later motivate \tether{} (Section~\ref{sec:tether}), which combines the complementary benefits of LOO and learned value-function baselines.
\paragraph{Miscellaneous.}
A \pvf{} can also condition on verifier rubrics or other detailed task specifications unavailable to the policy. Future work could explore \emph{reasoning value functions} that generate a chain of thought before predicting the value. Their additional inference-time compute could enable richer forms of privileged context, such as access to tools unavailable to the policy itself.

\subsection{Relation to self-distillation}

On-policy self-distillation methods similarly exploit privileged training-time information to provide dense token-level supervision rather than rely on sparse terminal rewards \citep{hubotter2026sdpo,zhao2026opsd}. For example, SDPO conditions the current model on retrospective environment feedback to generate the teacher distribution, and trains the student to match this teacher. Self-distillation introduces a new distribution-matching objective, typically through a token-level reverse KL term toward the conditional teacher policy, and therefore changes the policy optimum. This requires careful control over the information exposed to the teacher. If the teacher relies too strongly on privileged context, it may assign high probability mass to reasoning that the student cannot reproduce from its own observations. The teacher distribution can also become overly concentrated or simply drift too far from the student, making optimization unstable. Several prior self-distillation methods have studied additional stabilization mechanisms, information control and careful hyperparameter tuning to match GRPO baselines \citep{penaloza2026privilegedinformationdistillationlanguage, xu2026betaopsdderivingpolicyoptimization, zhu2026facesonpolicydistillationpitfalls}.

A \pvf{} by contrast, uses privileged information only to condition the value model which is an auxiliary tool for optimization. Under the admissibility condition in Equation~\ref{eq:baseline-identity} and GAE $\lambda_{\mathrm{GAE}}=1.0$, it can improve gradient variance without changing its expectation or the optimal policy under the original RL objective. Even with $\lambda < 1$ a perfectly fit value function maintains an unbiased policy gradient. In practice, this makes privileged information considerably easier to incorporate and tune with \pvf{}s. One tradeoff is that \pvf{}s cannot incorporate feedback produced from the completed response itself. Self-distillation does not share this restriction and can directly exploit retrospective critiques, verifier feedback, or other signals that depend on future states from the trajectory.

\section{Privileged value function experiments}
\label{sec:pvf-experiments}


\begin{figure}[t]
  \centering
  \includegraphics[width=\linewidth]{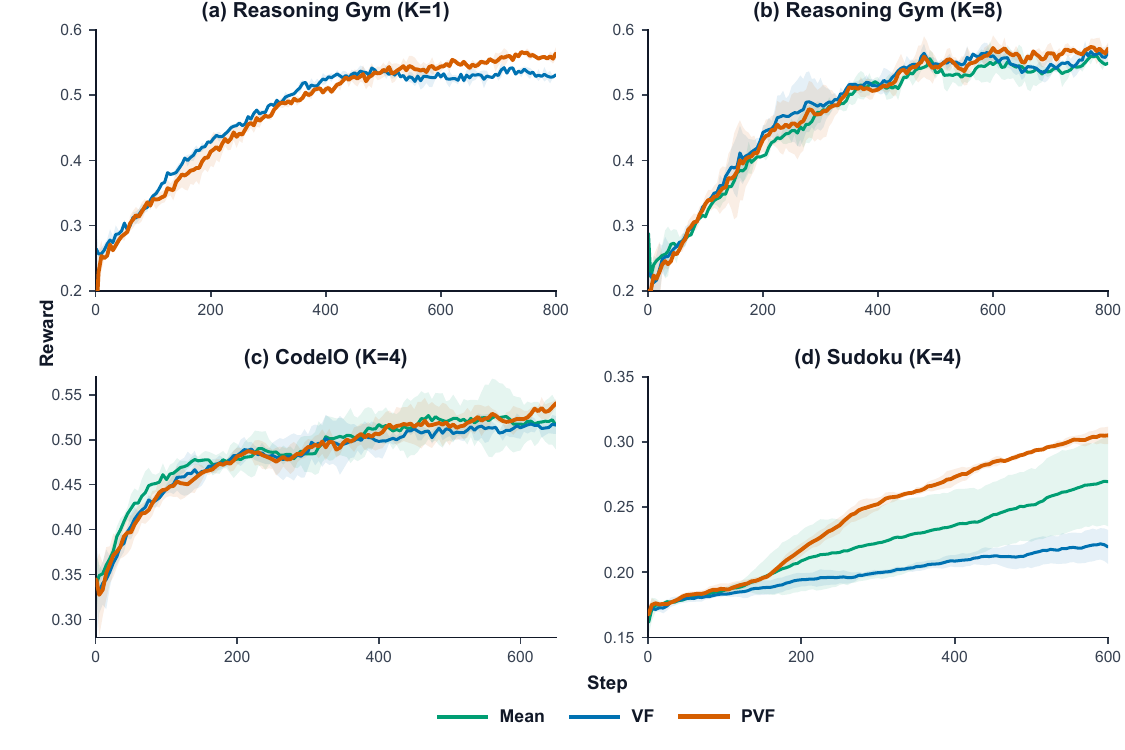}
  \caption{\textbf{RL with privileged value functions.} Seed-averaged training
  reward curves (EMA smoothed); shaded regions denote one
  standard deviation across seeds. 2 seeds for RG runs and 3 seeds for CodeIO and
  Sudoku. We compare
  the group-mean (\textsc{Mean}), ordinary value function (\textsc{VF}), and
  privileged value function (\pvf{}) baselines.  $K$ is the rollout group size. The no-group $K=1$ RG
  setting skips mean baseline. In RG and Sudoku, the
  privileged critic receives the ground-truth answer. In CodeIO, it receives the
  other $K-1=3$ leave-one-out responses from the group and their rewards. \pvf{} is the best performing method in all settings. Figure~\ref{fig:aggregate-final-window} summarizes end-of-training rewards.}
  \label{fig:pvf-main-results}
\end{figure}

\subsection{Experimental setup}
\label{sec:pvf-exps}

We compare three baselines. \textsc{Mean} is the group-mean GRPO baseline from
Equation~\ref{eq:group-mean}; \textsc{VF} is the ordinary token-level value
baseline from Equation~\ref{eq:mc-value}; and \pvf{} uses the same value
training configuration as \textsc{VF}, but conditions on task-specific privileged
context as in Equation~\ref{eq:privileged-value}. Both value-based methods use Monte Carlo targets and advantages, with
$\lambda_{\mathrm{target}}=\lambda_{\mathrm{GAE}}=1$. Within each
environment, policy training settings are matched across baselines, and all
runs use \texttt{Qwen3-4B-Instruct-2507} \citep{qwen3technicalreport}. Value functions are trained with the asynchronous infrastructure described in
Appendix~\ref{sec:infrastructure}. Experiment settings are detailed in Appendix~\ref{app:experimental-settings}.

\paragraph{Tasks.}
We launch four experiments across three environments:
\begin{itemize}
  \setlength{\itemsep}{0.15em}
  \setlength{\parsep}{0pt}
  \setlength{\topsep}{0.3em}
  \setlength{\partopsep}{0pt}
  \item \textbf{Reasoning Gym \citep{stojanovski2025reasoninggymreasoningenvironments}:} a weighted mixture of procedural,
  single-turn reasoning tasks from the Reasoning Gym suite. The \pvf{} receives
  the ground-truth answer. We evaluate both the no-group setting $K=1$ and a
  grouped setting with $K=8$. We set batch size as 128 and total sequence length as 8192 (prompt + response).
  \item \textbf{CodeIO \citep{li2025codeio}:} a single-turn program-reasoning task in which the
  policy predicts an output from an input, or a feasible input from an output.
  We test Leave-One-Out group info for this task, the \pvf{} receives the
  other $K-1=3$ responses in the rollout group and their returns. We set batch size as 128 and response length 8192.
  \item \textbf{Sudoku:} it is set up as a multi-turn environment where the policy reasons
  and fills one missing grid cell per turn. The \pvf{} receives the complete
  solved grid, while the policy observes only the standard interaction. The batch size is 64, total response length is 32768, and group size $K=4$.
\end{itemize}

\subsection{Results}

Figure~\ref{fig:pvf-main-results} and Figure~\ref{fig:aggregate-final-window} show that privileged conditioning improves
the value baseline across all four tasks, although the gains vary substantially. Since these experiments use Monte Carlo advantages, we highlight that the privileged signal affects policy learning only through baseline variance reduction.

In Reasoning Gym, \textsc{VF} and \pvf{} improve at similar rates through much
of training in the $K=1$ comparison, but \textsc{VF} plateaus earlier. The
improvement is smaller but still present at $K=8$, and both value baselines beat
\textsc{Mean} in this task. The CodeIO result is notable because the \pvf{} uses
no task-specific information, conditioning instead on other responses in the
rollout group and their returns. While \textsc{VF} slightly underperforms
\textsc{Mean}, \pvf{} surpasses both baselines, with its gap widening over the
course of training. We see the largest improvement from privileged conditioning
in Sudoku, which we hypothesize is partly due to the long, multi-turn horizon. Evaluating
an intermediate move requires knowing whether the partial grid is compatible with a globally consistent solution. An ordinary value function
must infer this implicitly, effectively solving much of the puzzle as
part of value prediction. Access to the solved grid removes this latent
inference problem and greatly simplifies the critic's task.

\subsection{\pvf{} better explains return variance}

Explained variance (EV) measures how much of the observed variation in rewards is
captured by the value predictions. For each value batch $\mathcal B$, we
compute
\begin{equation*}
  \widehat{\mathrm{EV}}
  =1-\frac{\Var_{\mathcal B}(R_i-\widehat V_{i,t})}
  {\Var_{\mathcal B}(R_i)}.
\end{equation*}

\noindent
\begin{minipage}[t]{0.47\textwidth}
  \vspace{0pt}
  A value of one indicates perfect prediction, while zero means that the critic
  explains no more return variance than a constant baseline. With $\lambda_{\mathrm{GAE}}=1$, $R_i-\widehat V_{i,t}$ is the policy advantage, so EV is a direct measure of the advantage variance reduction provided
  by the critic. Figure~\ref{fig:pvf-explained-variance} shows that \pvf{} explains more return
variance than \textsc{VF} in every environment. The EV improvement is correlated with the final reward gaps in Figure~\ref{fig:aggregate-final-window}:
the explained-variance gap is smallest for Reasoning Gym at $K=8$, where the
reward difference is also smallest, and largest for Sudoku, where privileged
conditioning produces the largest reward improvement.
\end{minipage}
\hfill
\begin{minipage}[t]{0.50\textwidth}
  \vspace{0pt}
  \includegraphics[width=\linewidth]{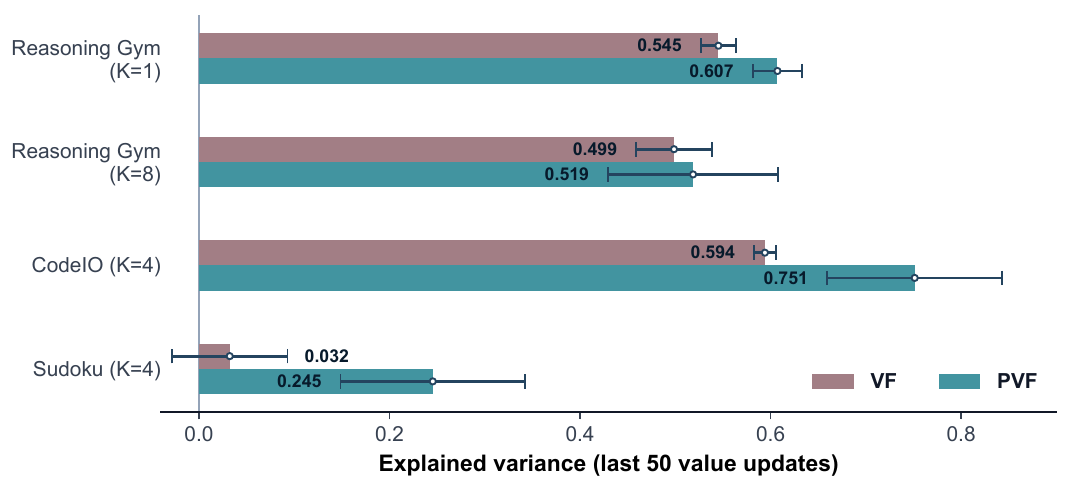}
  \refstepcounter{figure}
  \label{fig:pvf-explained-variance}
  {\small \textbf{Figure~\thefigure:} Explained variance of \textsc{VF} and
  \pvf{}, averaged across seeds and the final 50 value
  batches. Error bars show one standard deviation across seed means.}
\end{minipage}
\par\medskip

\subsection{Effect of group size on variance reduction}

Our experiments use relatively small group sizes, $K\in\{4,8\}$ accounting for our smaller batch sizes of 64 and 128. This raises a natural question: would substantially larger groups preferentially improve \textsc{Mean} relative to value function baselines? We reason why this likely would not be the case, but we welcome future work to investigate this. Increasing $K$ reduces variance through two distinct mechanisms, only one of which is specific to \textsc{Mean}. First, a larger group provides a more accurate estimate of the prompt-level baseline. For binary rewards with task success probability $p$, the variance of the LOO mean estimator is $p(1-p)/(K-1)$, arising purely from Bernoulli sampling noise. Even in the maximum variance case of $p=0.5$, doubling $K$ from 8 to 16 only reduces the standard baseline error from $0.19$ to $0.13$, while halving the number of unique tasks represented in a batch. Moreover, improving the prompt-level baseline may not reduce variance beyond the first token.

The second effect of increasing $K$ is obtaining more trajectories for each task and averaging their policy gradient contributions. This can continue to reduce within-task gradient variance even after baseline error is minimized, but it is not specific to \textsc{Mean} since value function methods using task groups (like in our experiments) receive the same benefit. The more critical limitation of value functions arises when the critic is poorly trained. In the next section, we discuss a baseline that adaptively trades off the benefits of value and mean baselines.

\section{\tether: A group-aware value baseline}
\label{sec:tether}

Value baselines only reduce variance better than the group mean when the critic is fit well. This can especially be problematic early in training when the value function has seen little data, assuming no extensive value pretraining phase. We might therefore like the baseline to begin near
the group mean and move toward token-level values as the critic improves during the course of RL. In doing so, we would also like to avoid any task specific hyperparameter tuning to manage this transition. We propose \tether, a mixture baseline which auto-interpolates between the mean and value baselines.

\subsection{Interpolating group and value baselines}

\tether{} is an adaptive linear combination of the mean and value baselines.
Let $V_{i,t}=V_\phi(h_{i,t})$ be the learned token value and let
$b_i^{\loo}$ be the leave-one-out group baseline from Equation~\ref{eq:loo}.
We define
\begin{equation}
  b^{\tether}_{i,t}
  =(1-\rho)b_i^{\loo}+\rho V_{i,t},
  \qquad
  A^{\tether}_{i,t}=R_i-b^{\tether}_{i,t}.
  \label{eq:tether}
\end{equation}
At $\rho=0$, \tether{} recovers the leave-one-out group advantage; at
$\rho=1$, it recovers the token-level value advantage. Intermediate values of $\rho$ add
token-level variation from the critic while dampening its prediction errors
with the group component.

A fixed $\rho$ assumes that the relative quality of the two estimates remains
stable. In practice, the value function usually improves as it receives more
updates, but can also temporarily fall behind a shifting policy. We therefore
fit the mixture $b^{\tether}(\rho)$ in the same way as a value function: choose the mixture ratio which best predicts the observed return-to-go. With terminal returns the target for every token prefix is simply $R_i$. Let $\mathcal B_k$ denote the current batch and $\rho_{k-1}$ the smoothed coefficient available at the start of the step. \textbf{A batch never uses a mixture coefficient fitted on its own returns.} We first compute and freeze the advantages used for policy training with $\mathcal B_k$ using $\rho_{k-1}$. Only afterward do we use the returns in $\mathcal B_k$ to fit
\begin{equation}
  \widehat\rho_k
  =\arg\min_{\rho}
    \sum_{(i,t)\in\mathcal B_k}
    \left(R_i-b^{\tether}_{i,t}(\rho)\right)^2,
  \label{eq:tether-regression}
\end{equation}
The resulting batch estimate $\widehat\rho_k$ is a measure of how much
the learned value improves return prediction over the group baseline. The least squares optimization is very computationally cheap and can be easily done every step. Since the minimizer of any finite batch is noisy, we further reduce variance by
EMA smoothing successive estimates,
\begin{equation}
  \rho_k=d\rho_{k-1}+(1-d)\widehat\rho_k,
  \label{eq:tether-ema}
\end{equation}
where $d$ is the EMA decay. We emphasize again that the updated coefficient $\rho_k$ is used to compute advantages for
the next batch, $\mathcal B_{k+1}$
and not $\mathcal B_k$. Not doing so would make the baseline for $\mathcal B_k$ depend on
its own trajectory returns, violating the condition in
Equation~\ref{eq:baseline-identity} and biasing the policy gradient. Initializing training with $\rho=0$ starts
from the group baseline, and with good value training we should expect the coefficient to move towards $\rho=1$ as
the critic accuracy improves. Appendix~\ref{app:tether-variance} provides additional statistical analysis of the \tether{} baseline.

\subsection{\tether{} as a privileged value function}

\tether{} can also be viewed as a simple \pvf{}. The
sibling returns $\{R_j\}_{j\ne i}$ are privileged information for trajectory
$i$, which the leave-one-out mean reduces into the scalar $b_i^{\loo}$.
Equation~\ref{eq:tether} then linearly interpolates this group-conditioned estimate and the
ordinary token value, with only the mixture coefficient
$\rho$ ``learned'' (least-squares fit) from data. This makes it clear why we need to use the LOO estimate and not full group mean to satisfy the unbiasedness condition in Equation~\ref{eq:baseline-identity}.

The leave-one-out \pvf{} used in the CodeIO experiments in Section~\ref{sec:pvf-exps} is a more expressive realization of the same idea. It jointly conditions on the current trajectory's partial chain of thought and the complete LOO group responses and rewards, allowing the critic to flexibly learn which group information is relevant to the current state's value. \tether{}, by contrast, first mean reduces the LOO returns and then combines this scalar with the ordinary token value through linear combination.

\section{\tether{} experiments}
\label{sec:tether-experiments}

\subsection{Experimental setup}

We compare \tether{} against the same \textsc{Mean} and \textsc{VF} baselines used in
Section~\ref{sec:pvf-experiments}. We set EMA decay $d=0.95$. The value-function training
configuration is shared between \textsc{VF} and \tether{}. The task suite is the same as in Section~\ref{sec:pvf-experiments}, with two changes: 1) we omit the $K=1$ Reasoning Gym experiment because \tether{} requires group
rollouts, 2) we add MiniF2F \citep{zheng2022minif2fcrosssystembenchmarkformal}, a multi-turn Lean formal mathematics task. The
policy receives compiler feedback after each turn for up to 3 attempts, with up to 4096 generated tokens per attempt. We use \texttt{Qwen3.5-4B} \citep{qwen3.5} for this task, as \texttt{Qwen3-4B-Instruct-2507} failed to obtain any training signal. Other tasks use \texttt{Qwen3-4B-Instruct-2507} as before. Detailed experiment settings in Appendix~\ref{app:experimental-settings}.

\begin{figure}[!h]
  \centering
  \includegraphics[width=\linewidth]{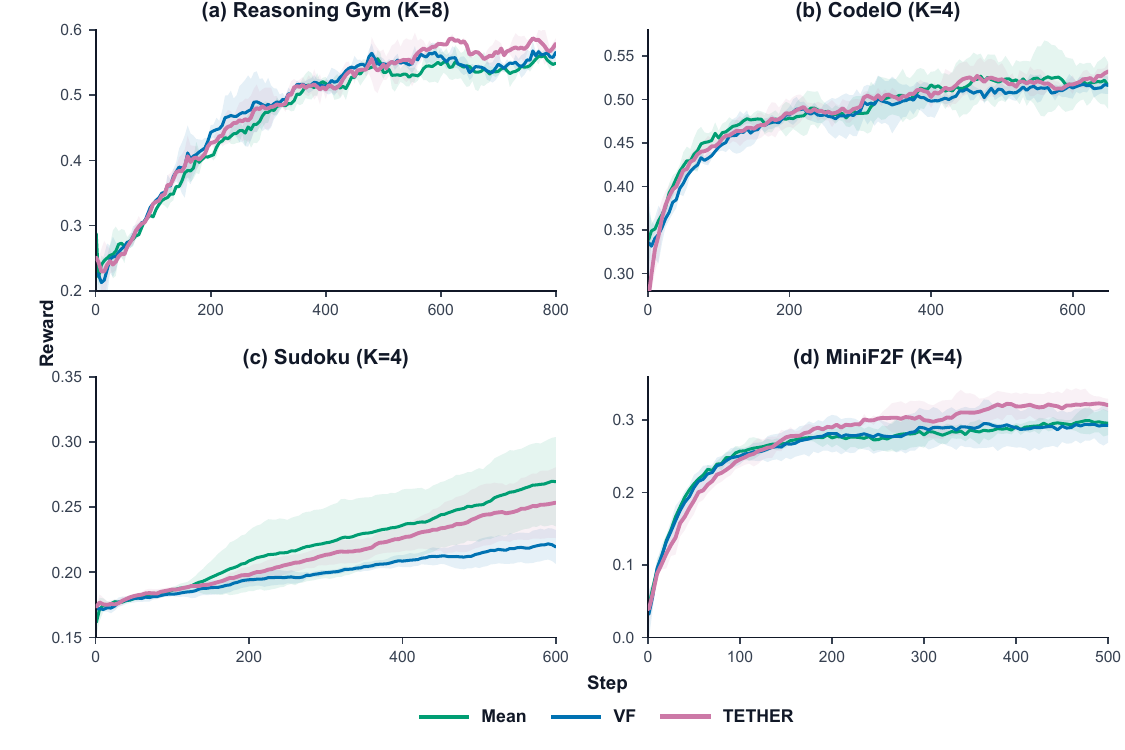}
  \caption{\textbf{RL with \tether{} baseline.} Seed-averaged
  training reward curves (EMA smoothed); shaded regions denote one standard
  deviation across seeds. Reasoning Gym uses two seeds, while CodeIO, Sudoku,
  and MiniF2F use three. We compare the group-mean (\textsc{Mean}), ordinary
  value-function (\textsc{VF}), and adaptive \tether{} baselines. $K$ denotes
  the rollout group size. \tether{} improves over \textsc{VF} in all four
  settings. \tether{} outperforms \textsc{Mean} in RG and MiniF2F, matches it in CodeIO, and shrinks the gap in Sudoku. Figure~\ref{fig:aggregate-final-window} summarizes end-of-training rewards.}
  \label{fig:tether-main-results}
\end{figure}

\subsection{Results}

Figure~\ref{fig:tether-main-results} and Figure~\ref{fig:aggregate-final-window} show that \tether{} consistently outperforms the simple value function baseline (\textsc{VF}) across all four tasks. Although it does not fully recover the performance of \textsc{Mean} on Sudoku, it substantially mitigates the degradation suffered by \textsc{VF}. These results support our intuition that \tether{} is a reliable way to introduce value functions into GRPO pipelines that already perform well with the standard mean baseline. 

\subsection{Adaptive coefficient dynamics}

\noindent
\begin{minipage}[t]{0.48\textwidth}
  \vspace{0pt}
  Figure~\ref{fig:tether-rho} shows how the mixture coefficient $\rho$ evolves
during each experiment. All runs begin with $\rho=0$, so early policy updates
favor the LOO baseline. The coefficient then moves away from zero as the critic
begins to explain return variance beyond the group mean. Interestingly, the $\rho$ convergence value is strongly
  task-dependent, and Sudoku converges to the largest value-function mixture
  weight despite exhibiting the weakest \textsc{VF} baseline performance. We
  hypothesize that relying exclusively on a poorly fitted value function early
  in training impedes initial policy learning, producing compounding effects
  that slow subsequent progress. \tether{} mitigates this failure mode by
  relying on the more dependable group baseline initially.
\end{minipage}
\hfill
\begin{minipage}[t]{0.49\textwidth}
  \vspace{0pt}
  \includegraphics[width=\linewidth]{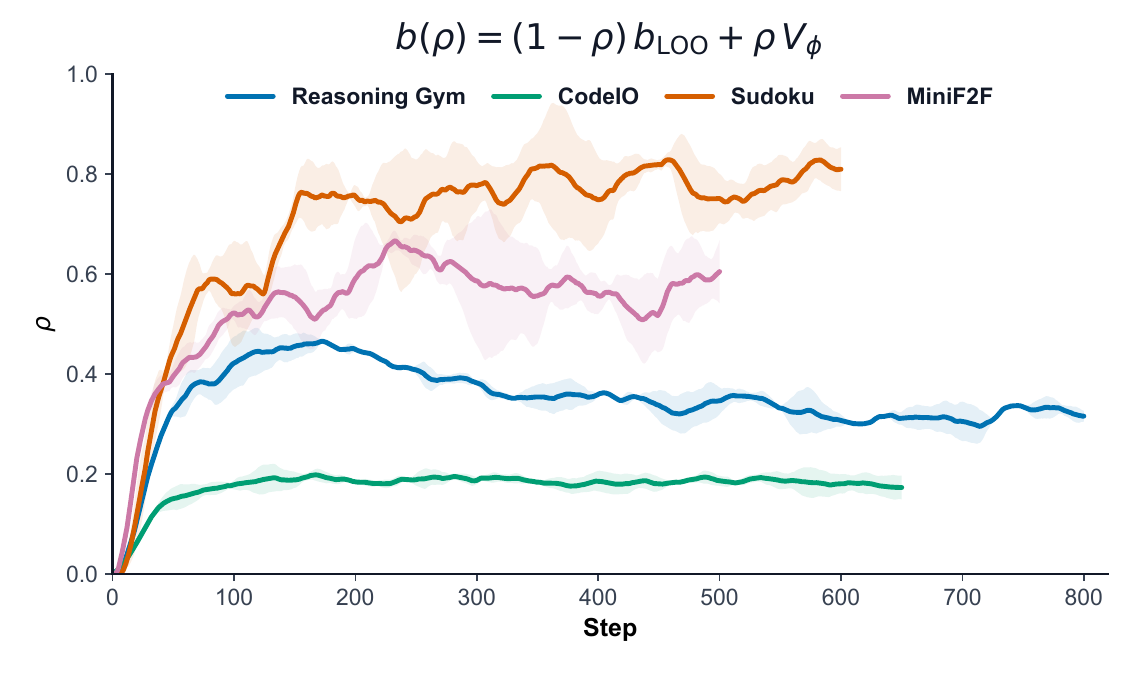}
  \refstepcounter{figure}
  \label{fig:tether-rho}

  {\small \textbf{Figure~\thefigure:} Seed-averaged \tether{} coefficient
  $\rho$, adaptively fit during RL. Values closer to $\rho=1$ indicate that
  the value function predicts return-to-go better than the mean
  baseline.}
\end{minipage}
\par\medskip

\begin{figure}[!h]
  \centering
  \includegraphics[width=\linewidth]{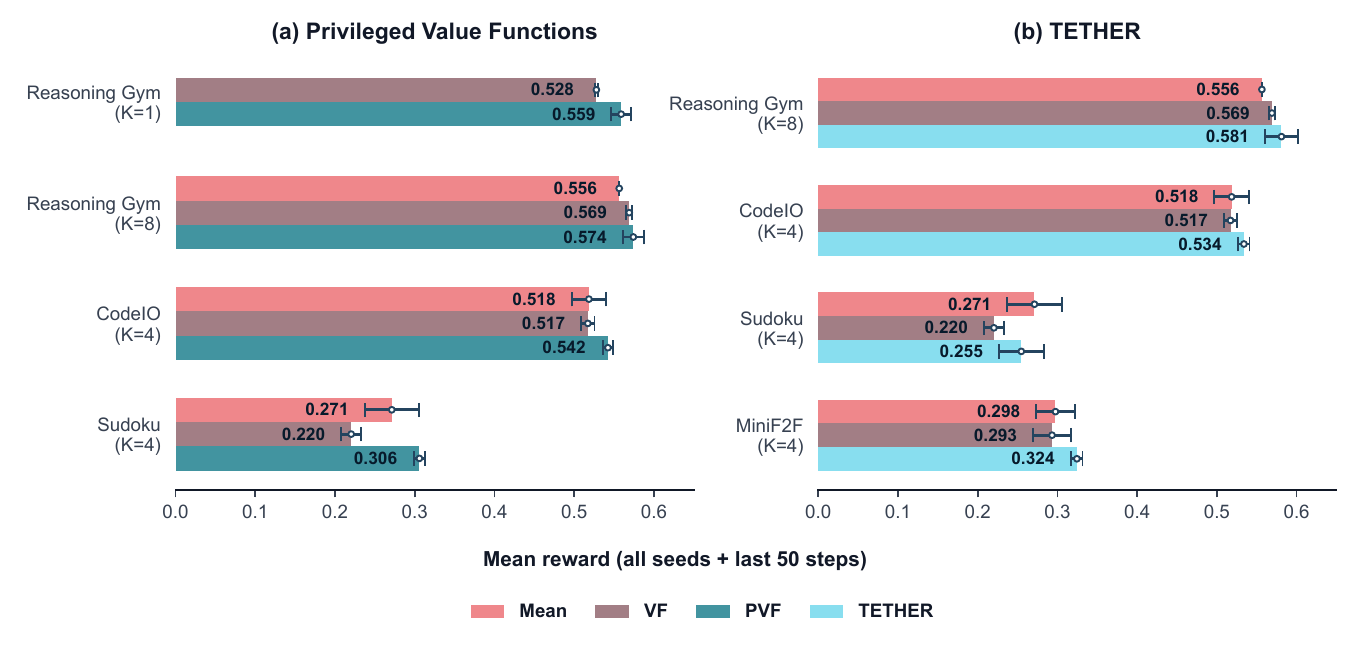}
  \caption{\textbf{Aggregated final results.} Bars show the mean raw
  reward over the final 50 policy steps, first averaged within each seed and
  then across seeds for every run; error bars denote one standard deviation across the
  per-seed means. This summary helps directly compare training reward differences between methods. Left: \pvf{} experiments. Right: \tether{} experiments.}
  \label{fig:aggregate-final-window}
\end{figure}

\section{Potential value function research for future work}
\label{sec:future}
An objective of this paper is to motivate the research community to reconsider the utility of value functions for LLM RL, and in this spirit, we now discuss some promising directions for future work. We believe that the privileged value techniques proposed in this paper, combined with further engineering optimization, could make value functions a practical and scalable component of large-scale LLM post-training.

\subsection{Extensive value pretraining from diverse policies}
In all our experiments, we initialized the value function as a copy of the base policy with a randomly initialized value head and used only 20 value pretraining steps before starting RL. This was intended to roughly match the number of inference trajectories between the mean and value baseline experiments for a straightforward comparison. Even with this simple init, our experiments show that value functions can outperform the mean baseline when they benefit from useful privileged information. Several prior works have found that value functions benefit significantly from extended pretraining \citep{yuan2025vcppo, hou2026singleasync, yue2025vapo}. We further posit that it may be beneficial to do value pretraining using data generated by diverse policies rather than only the static base policy. This could improve the value function's adaptability as the policy shifts during RL and reduce overfitting to the initial policy. This idea has recently been validated by  \cite{dong2026reallyneedpretrainqfunctions} for non-LLM RL. The pretraining dataset should include samples containing privileged information if we want \pvf{}s. Pretrained value functions could potentially be reused across many RL runs, amortizing training cost.

\subsection{Tuning $\lambda$ for bias--variance trade-off}

Both $\lambda_{\mathrm{GAE}}$ and $\lambda_{\mathrm{target}}$ (defined in
Section~\ref{sec:value-summary}) are highly sensitive value-function
hyperparameters. Figure~\ref{fig:lambda-ablation} compares
$\lambda_{\mathrm{GAE}}=1$ and $0.999888$ on Reasoning Gym. The latter $\lambda$ was chosen such that the first token advantage retains approximately $40\%$ of the unbiased terminal signal at 8192 length response, i.e. $\lambda=0.4^{1/8192}\approx0.999888$. Despite their small magnitude difference, the lower $\lambda_{\mathrm{GAE}}$ significantly improves both \textsc{VF} and
\pvf{} rewards. We still used $\lambda_{\mathrm{GAE}}=\lambda_{\mathrm{target}}=1.0$ for all our main experiments in Sections~\ref{sec:pvf-experiments} and \ref{sec:tether-experiments} to focus our primary analysis on the unbiased advantage setting. This experiment indicates that coarse $\lambda$ sweeps can easily miss this narrow but consequential regime near $\lambda=1$.

\begingroup
\raggedbottom
\newpage
\endgroup
\noindent
\begin{minipage}[t]{0.47\textwidth}
  \vspace{0pt}
    Several prior studies compare $\lambda=1$ against substantially smaller values which drown out the terminal reward signal at their sequence lengths
  \citep{ahmadian2024rloo,kazemnejad2025vineppo,yuan2025vcppo,
  hu2025openreasonerzero}. DeepSeek-R1, generally credited for popularizing critic-free RL, reports a comparison only between
  $0.95$ and $1.0$ for long sequence RL \citep{deepseekai2025r1}. Some recent work calibrates
$\lambda$ to the sequence length which we expect to generally work better due to the exponential terminal reward decay \citep{yue2025vapo,hou2026singleasync}. Future
work could study $\lambda$-tuning systematically, including disentangling the effects of $\lambda_{\mathrm{GAE}}$ and $\lambda_{\mathrm{target}}$, and exploring techniques for adaptive $\lambda$-tuning as a function of critic accuracy.
\end{minipage}
\hfill
\begin{minipage}[t]{0.50\textwidth}
  \vspace{0pt}
  \includegraphics[width=\linewidth]{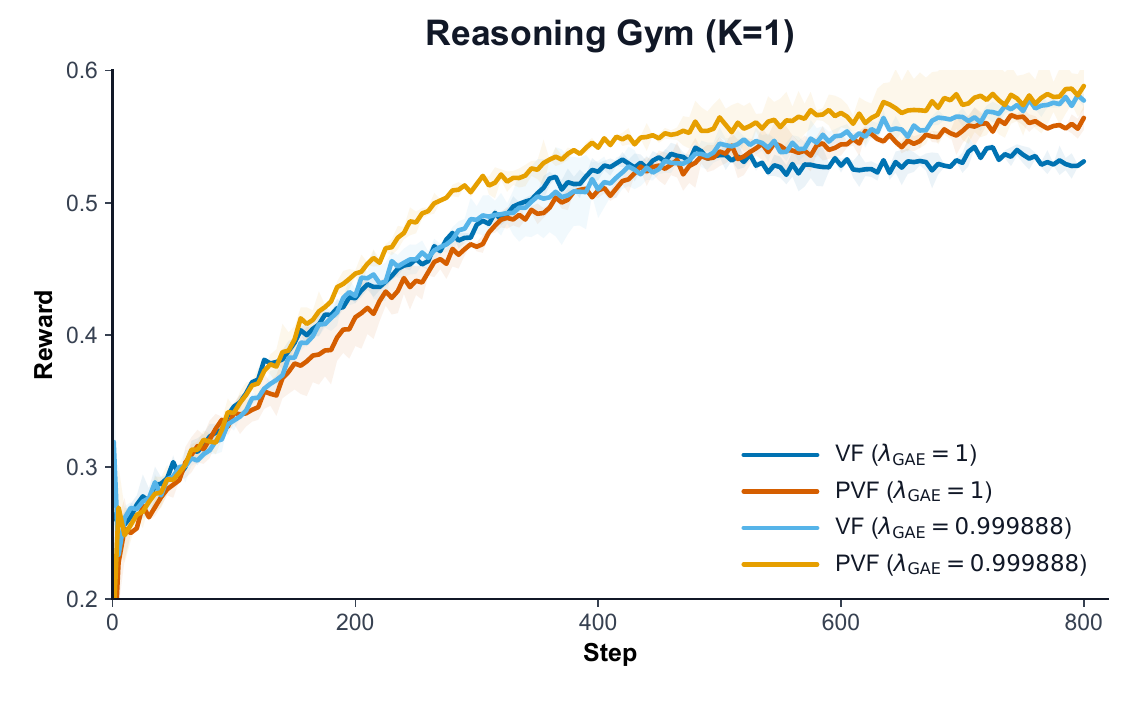}
  \refstepcounter{figure}
  \label{fig:lambda-ablation}

  {\small \textbf{Figure~\thefigure:} Policy reward for $\lambda_{\mathrm{GAE}}\in\{0.99988,1.0\}$ with ordinary (\textsc{VF}) and privileged (\pvf) value functions. Curves are 2-seed-averaged and
  EMA-smoothed.}
\end{minipage}
\par\medskip

\subsection{Token-bucketed \tether}
In Section~\ref{sec:tether}, we fit a single coefficient $\rho$ to mix the
group and value baselines. This assumes that the
relative quality of the two baselines is uniform across all token positions in the response, which is not generally true. For example, in Sudoku, value prediction may become substantially easier later in a trajectory. When the grid is nearly filled, the critic only needs validate it, whereas earlier it must implicitly marginalize over many possible completions.
 A simple heuristic accounting for this is to use different mixing coefficients depending on token position. We can partition response tokens into $M$ buckets (say uniformly binned from $0$ to max response length) and then fit a separate coefficient $\rho_m$ for
each bucket. If $m(t)$ denotes the bucket containing token position $t$:
\begin{equation}
  b^{\tether}_{i,t}
  =\left(1-\rho_{m(t)}\right)b_i^{\loo}
   +\rho_{m(t)}V_{i,t}.
\end{equation}
Each $\rho_m$ can be fit using the same return prediction objective as
Equation~\ref{eq:tether-regression} restricted to tokens that fall in its bucket, and smoothed with its own EMA. We expect the bucket size would need to be carefully considered, since bucket sizes too small would have fewer token samples to fit $\rho$, thus making estimation noisier.

\section{Limitations}

\paragraph{Value functions add infra cost.}
Value function inference and training add accelerator cost to the workload, requiring dedicated GPU allocation. We do not exactly compute-match \textsc{Mean} and value function baselines in our experiments, instead matching only the number of inference trajectories. However, the \textsc{VF} and \pvf{} settings are exactly matched. Specific compute details are provided in Appendix~\ref{app:experimental-settings}. More generally, we would like future work to investigate \textit{value function scaling laws}, optimizing value function compute allocation at different compute scales, and comparing scaling trends against critic-free baselines.

\paragraph{Small-scale experiments.}
Our experiments are limited to 4B models, on tasks with maximum response length of 32,000 tokens (Sudoku). We would like to see experiments extended to long-horizon agentic training settings, where we expect the variance reduction provided by value functions to be even more pronounced.

\section{Conclusion}
\label{sec:conclusion}

Value functions have substantial untapped utility for LLM RL. Beyond variance reduction considered in this paper, they could provide non-terminal learning signals for partial trajectories, enabling policy updates before expensive long-horizon episodes have finished sampling. This could be particularly desirable as agents are trained for longer horizon tasks. They may also be used for inference-time scaling. In this work, we improved value functions as control variates by conditioning them on privileged information, offering a reliable alternative to self-distillation. We also introduced the \tether{} baseline, which provides a natural interpolation between group-relative and value-based RL and therefore a low-risk path to integrate value functions with existing GRPO infra. We hope these methods motivate further research into effective use of value functions.


\bibliographystyle{plainnat}
\bibliography{references}

\clearpage
\appendix
\section{Related work}
\label{app:related-work}

\paragraph{Privileged conditioning of value functions.}
Outside LLM post-training, there have been several asymmetric actor--critic implementations that have allowed the critic to observe
training-time privileged information hidden from the policy. \citet{pinto2017asymmetric} used the
full simulator state in the critic to train visual-control policies. \citet{baisero2022unbiased} analyzed when privileged-information critics remain unbiased under partial observability conditions. \citet{hu2024privilegedsensingscaffoldsreinforcement} show that an asymmetric critic can eliminate error terms caused by aliasing in the agent's observable state, providing a theoretical explanation for improved convergence from privileged information. These prior works in general applied the idea to simulated robotics control. The closest LLM variant we are aware of is SWEET-RL \citep{zhou2025sweetrl} which also uses privileged information to construct advantages, although there are several important differences from our work. Rather than training a token-level value function to predict the expected return, SWEET-RL learns a turn-level, action-conditioned advantage model from reward-ranked trajectory pairs using a Bradley-Terry objective. While our value model is then used as a baseline or with GAE for policy gradients, SWEET-RL instead uses their advantage model to construct preference pairs for DPO \citep{rafailov2023dpo}.

\paragraph{Combining group and value baselines.}
The closest concurrent work is
EVPO \citep{pan2026evpo} which computes critic explained return variance on each batch
and makes a hard choice to use the critic when explained variance is positive, and otherwise use the group mean. \tether{} instead smoothly interpolates between the group
baseline and token-level values directly for return prediction, which theoretically has some advantages, which we compare in Appendix~\ref{app:tether-variance}.

\paragraph{LLM credit-assignment without critics.}
VinePPO  \citep{kazemnejad2025vineppo} decides to forego critic learning, instead estimating intermediate values by branching Monte-Carlo rollouts from token
prefixes. A separate line of on-policy self-distillation
conditions a teacher on completed trajectories, critiques, or other
training-time information and trains the policy to match the resulting distribution
\citep{zhao2026opsd,hubotter2026sdpo,penaloza2026privilegedinformationdistillationlanguage,xu2026betaopsdderivingpolicyoptimization}.
Such methods can use retrospective information that is inadmissible for an
unbiased baseline, but they introduce a biased policy distillation objective.

\section{Asynchronous value function training infrastructure}
\label{sec:infrastructure}

Our value function implementation is open sourced as
\href{https://github.com/HyperPotatoNeo/prime-values}{\texttt{prime-values}},
built on top of PRIME-RL \citep{primeintellect2025primerl}. The high level design goal is to keep value learning from becoming a synchronization barrier for the RL loop, and support replay buffer training. Policy updates are sensitive to off-policy lag requiring a balanced inference-trainer pipeline for maximum throughput. Value training on the other hand is generally more tolerant of
stale data, and trajectories can be reused for several updates. The policy distribution
shift eventually makes sufficiently old data problematic for value training, but the acceptable
window is generally wider than for policy updates which must be importance ratio corrected (which increases variance). Our system exploits this
asymmetry while still bounding both the age and reuse of critic data through configuration of the replay buffer. We encourage future work to use our infrastructure to systematically study optimal compute and data allocation for value function training.

\subsection{Value evaluation and training}

The value function system takes on two logical roles. The \emph{value trainer}
performs optimizer updates and publishes monotonically versioned weights. The
serving copy, called the \emph{value evaluator} in our implementation, predicts
the expected return from each partial response, used to compute advantages for policy training. The value trainer feeds on samples from a replay buffer, and trajectories are only added to this buffer after advantages are computed using the value evaluator. This
evaluate-before-train configuration is necessary to maintain unbiased advantage estimation.

Our infra supports two modes of compute placement for these roles. In the default \emph{dedicated}
placement, value training and evaluation use separate model replicas, usually
on separate GPUs. The evaluator adopts newly published trainer weights while
continuing to serve requests. Value inference can therefore overlap value training, so a busy trainer does not delay advantage construction (which would in-turn bottleneck policy training). The
cost is an additional serving allocation to store the extra value copy. In the \emph{colocated} placement, the trainer's GPUs also serve value
inference. This avoids the dedicated evaluator allocation, but the same model
cannot train and serve simultaneously. The runtime alternates complete
optimizer steps with complete value inference batches. Every inference batch takes
time away from value optimization, reducing the number of critic updates that
fit into a run. Conversely, an advantage request arriving during a long optimizer step
must wait for that step to finish. This delays advantage computation for an
otherwise ready trajectory and can increase the time between generation and
its eventual policy update, potentially increasing off-policyness.

\subsection{FIFO-bounded replay and controlled reuse}

Complete trajectories which have finished value evaluation for advantage estimation are pushed into the FIFO replay buffer. Trajectories are sampled in batches (without replacement) uniformly from this replay buffer for value training. Each trajectory has associated with it a replay counter that is incremented whenever it is sampled to a training batch. We evict trajectories once their replay counter exceeds the sample reuse limit, which we fix to $N=2$ for all experiments. We found training to remain stable for $N>2$, but our inference node generated trajectories quickly enough that even with $N=2$, the value trainer maintained high GPU utilization with no idle time.

\begin{figure*}[t]
  \centering
  \includegraphics[width=0.98\textwidth]{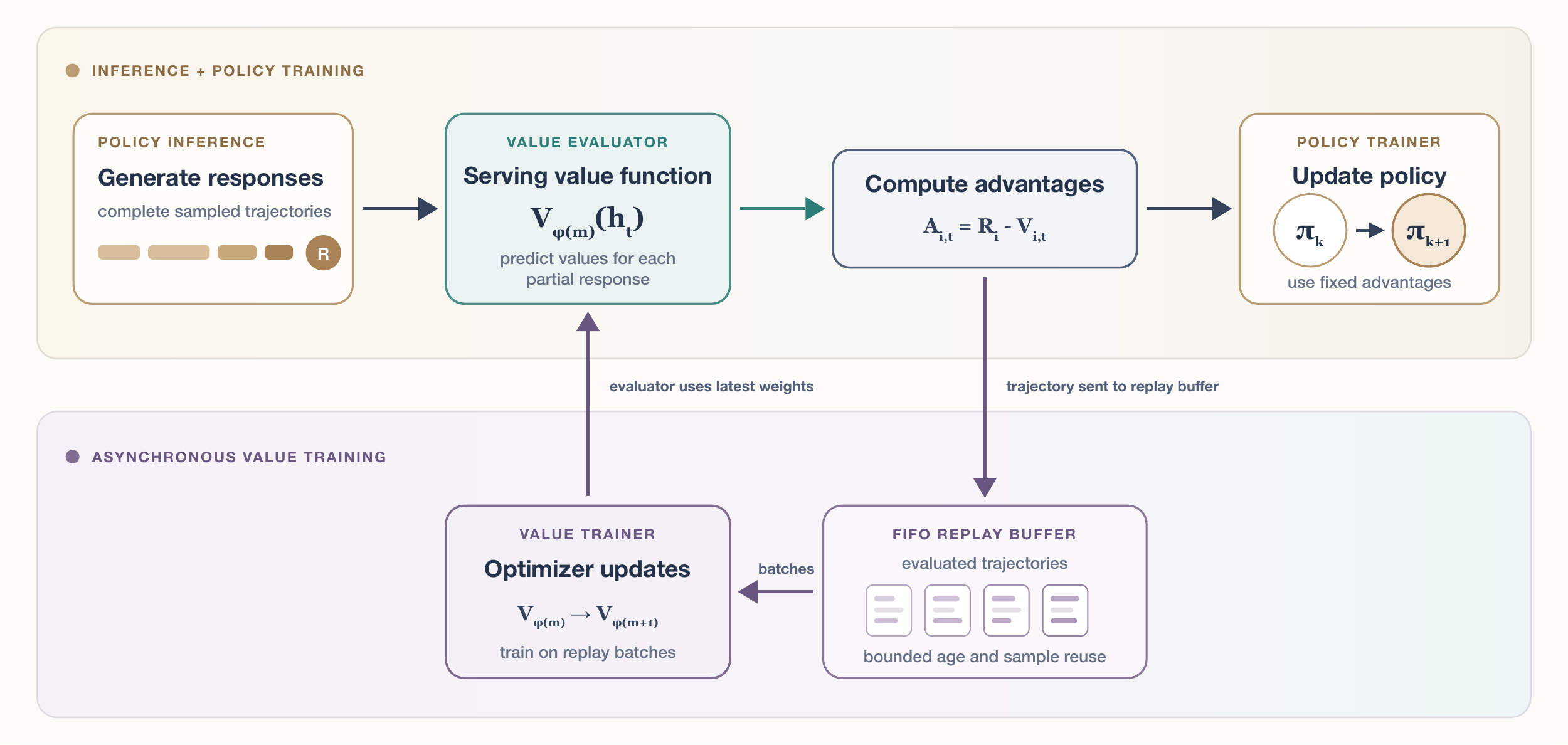}
  \caption{Asynchronous value function evaluation and training. The value evaluator is the serving
  copy used in the inference and policy-training pipeline: it predicts the
  expected return from every partial response and these predictions are used
  to compute policy advantages. Evaluated trajectories are then added to
  the FIFO replay buffer. Independently, the value trainer samples replay
  batches and performs optimizer updates; the evaluator uses the latest
  published weights.}
  \label{fig:async-value-architecture}
\end{figure*}

\subsection{Value warmup before policy training}

We initialize the value model as a copy of the base policy with a randomly
initialized value head. Before allowing policy updates, we run the same
asynchronous pipeline for a short value pretraining phase. The critic uses the same rollout batch size as in
the subsequent RL phase. In all our experiments we only use
20 value updates before beginning policy optimization, so this is more like value warmup than pretraining. This provides mostly a reasonable calibration for the trained baseline without requiring a separately generated value pretraining dataset. We expect substantially longer and more diverse value pretraining may be
beneficial, as discussed in Section~\ref{sec:future}.

\subsection{Value loss}

All our tasks use binary outcome rewards, and so we train the value head with a binary classification loss rather than MSE regression. Importantly, the predicted values are not binary, but rather the expectation of the bernoulli distribution $p$, and so is a continuous value in $[0, 1]$. Previous work in deep
RL without LLMs report that categorical value prediction can improve value training \citep{farebrother2024classification}. We observed a small
improvement over MSE in early experiments, however more experimentation would help to validate this decision. The effect of value loss and support warrants more systematic study.

\section{Statistical analysis of \tether{}}
\label{app:tether-variance}

For a sampled token, let $R$ be its trajectory's observed terminal return,
$B$ the leave-one-out group baseline, $V$ the token-level value prediction,
and $\rho\in[0,1]$ the weight placed on the value prediction. \tether{} uses
\begin{equation}
    b_\rho=(1-\rho)B+\rho V,
    \qquad
    A_\rho=R-b_\rho.
    \label{eq:tether-variance-setup}
\end{equation}
To make the role of $\rho$ explicit, define the group advantage
$A_B=R-B$ and the value correction $\Delta=V-B$. Then
$A_\rho=A_B-\rho\Delta$.

\paragraph{Why a soft mixture can help.}
Moving $\rho$ from zero to one moves the baseline from $B$ toward $V$.
\tether{} chooses the point on this line that best predicts the observed
Monte Carlo return, minimizing squared error. At the population level, this is
equivalently the second moment of the resulting advantage:
\begin{align}
    \mathcal{L}(\rho)
    &= \E\!\left[(R-b_\rho)^2\right]
     = \E[A_\rho^2] \\
    &= \E[A_B^2]
       -2\rho\E[A_B\Delta]
       +\rho^2\E[\Delta^2].
    \label{eq:tether-variance-quadratic}
\end{align}
When $\E[\Delta^2]>0$, the optimal mixture is
\begin{equation}
    \rho^\star
    =
    \clip_{[0,1]}\!\left(
        \frac{\E[A_B\Delta]}
             {\E[\Delta^2]}
    \right).
\end{equation}
The numerator measures whether moving from the group baseline toward the value
prediction reduces the current residual; the denominator accounts for how far
apart the two predictions are.

Because $\rho=0$ recovers the group baseline and $\rho=1$ recovers the value
baseline, optimizing over the full interval immediately gives
\begin{equation}
    \E[A_{\rho^\star}^2]
    \leq
    \min\!\left\{
        \E[(R-B)^2],
        \E[(R-V)^2]
    \right\}.
    \label{eq:tether-variance-dominance}
\end{equation}
The inequality is strict when the optimum lies inside $(0,1)$. Intuitively,
this occurs when the two baselines make complementary errors, so averaging them
predicts the return better than either alone. Figure~\ref{fig:tether-rho} shows that the fitted
coefficient typically remains comfortably in the interior of $[0,1]$ in real training runs, indicating that
\tether{} uses information from both baselines rather than collapsing to either
endpoint.

\paragraph{Relation to optimal policy gradient variance.}
Let $s=\nabla_\theta\log\pi_\theta(y_t\mid h_t)$ and
$g_\rho=A_\rho s$. When $\rho$ is fixed for the current update and $B$ and
$V$ are admissible baselines, $\E[g_\rho]$ does not depend on $\rho$, while
\begin{equation}
    \operatorname{tr}\operatorname{Cov}(g_\rho)
    =
    \E\!\left[A_\rho^2\lVert s\rVert^2\right]
    -
    \lVert\E[g_\rho]\rVert^2.
\end{equation}
\tether{} does not take into account the $\lVert s\rVert^2$ term, and therefore is not the optimal policy gradient baseline, which is usually impractical to compute.

\paragraph{Comparison with EVPO.}
EVPO computes
\begin{equation}
    \widehat{\mathrm{EV}}_{\mathcal B}
    =
    1-
    \frac{\Var_{\mathcal B}(R-V)}
         {\Var_{\mathcal B}(R)}
\end{equation}
and uses the critic when this quantity is positive; otherwise it uses the full
group mean \citep{pan2026evpo}. EVPO therefore makes a hard endpoint choice
using the centered residual variance. In contrast, \tether{} directly fits Monte
Carlo squared error and allows an interior mixture. For any fixed pair of
baselines $B$ and $V$, oracle \tether{} cannot have higher Monte Carlo
prediction error than the better endpoint because $[0,1]$ contains both
choices $\{0,1\}$, and an interior optimum strictly improves on both.

\paragraph{Practical tradeoff.}
\tether{} is most attractive when the baselines have complementary errors and
enough stable data is available to estimate an interior coefficient $\rho$.
EVPO can be preferable when the optimum is near an endpoint, batches are too
small to estimate a continuous coefficient reliably, or the critic changes
too quickly for \tether{}'s smoothed, lagged coefficient. In finite data the
fitted coefficient can also overfit, so the population guarantee need not hold
on every policy update.

\section{Experimental hyperparameters and settings}
\label{app:experimental-settings}

This section reports the settings used for the experiments in
Figures~\ref{fig:pvf-main-results} and~\ref{fig:tether-main-results}. All nodes consist of 8$\times$H200 GPUs. Within the scope of each task, the policy configuration is shared across baselines. The ordinary (\textsc{VF}) and privileged value
function (\pvf{}) baselines always share the same value-training configuration and compute
allocation, only differing by critic context.

\subsection{Privileged value function experiments}

\paragraph{Reasoning Gym.}
We train \texttt{Qwen3-4B-Instruct-2507} for 800 policy steps with batch size
128 and evaluate $K\in\{1,8\}$. Each trajectory has an 8,192-token total
sequence limit (prompt+completion) and a 6,144-token completion cap. The \pvf{} receives the
reference answer as privileged context. The $K=8$ mean run uses two nodes: one for policy inference
and one for policy training. Both \textsc{VF} and \pvf{} use four nodes, adding
one value-trainer node and one dedicated value-evaluator node. We average two seeds.

\paragraph{CodeIO.}
We train \texttt{Qwen3-4B-Instruct-2507} for 650 policy steps with batch size
128 and group size $K=4$. The trajectory allows up to 4,096 input tokens
and 8,192 completion tokens, for a 12,288-token sequence limit. The \pvf{} conditions on the other $K-1=3$ responses in the group together with their returns.
Mean runs use three nodes---two for policy inference and one for policy
training. \textsc{VF} and \pvf{} use five, adding one value trainer
and one dedicated value evaluator. We average three seeds.

\paragraph{Sudoku.}
We train \texttt{Qwen3-4B-Instruct-2507} for 600 policy steps on the 5,000
hard-puzzle dataset, using batch size 64 and group size $K=4$. The environment
is multi-turn and asks the policy to fill one missing cell at a time. A
trajectory is limited to 32,768 tokens, with at most 8,192 generated tokens in
any model turn. The \pvf{} receives the complete solved grid. Mean runs use
four nodes---three for policy inference and one for policy training. \textsc{VF} and \pvf{} use six, adding one value trainer and one dedicated
value evaluator. We average three seeds.

\subsection{\tether{} experiments}

All \tether{} runs use EMA decay $d=0.95$ for the fitted mixture coefficient.
As above, \textsc{VF} and \tether{} share the value configuration and compute
allocation. Reasoning Gym uses two seeds and the other tasks use three.

\paragraph{Reasoning Gym.}
We use \texttt{Qwen3-4B-Instruct-2507}, batch size 128, group size $K=8$, and
800 policy steps. The total sequence and completion limits are 8,192 and 6,144
tokens, respectively. The mean baseline uses one inference and
one policy-trainer node. \textsc{VF} and \tether{} add one value-trainer node
and one dedicated value-evaluator node, for four nodes in total.

\paragraph{CodeIO.}
We use \texttt{Qwen3-4B-Instruct-2507}, batch size 128, group size $K=4$, and
650 policy steps, with 4,096 input tokens and up to 8,192 generated tokens.
Mean runs use two policy-inference nodes and one policy-trainer node.
\textsc{VF} and \tether{} additionally use one value trainer and one dedicated
value evaluator, for five nodes in total.

\paragraph{Sudoku.}
We use \texttt{Qwen3-4B-Instruct-2507}, batch size 64, group size $K=4$, and
600 policy steps. Trajectories are limited to 32,768 tokens and each turn
output to 8,192 tokens. Mean runs use three policy-inference nodes and one
policy-trainer node. \textsc{VF} and \tether{} add one value trainer and one
dedicated value evaluator, for six nodes in total.

\paragraph{MiniF2F.}
We train \texttt{Qwen3.5-4B} for 500 policy steps with batch size 128 and group
size $K=4$. Each theorem permits three proof attempts with compiler feedback,
up to 4,096 generated tokens per attempt and 24,576 tokens over the full
trajectory. Mean runs use 4 total nodes, three policy-inference nodes and one policy-trainer
node. \textsc{VF} and \tether{} use six nodes, adding one value trainer and one
dedicated value evaluator.

\subsection{Common training hyperparameters}

Table~\ref{tab:policy-hparams} gives the policy and rollout settings held fixed
across baseline comparisons. We use PRIME-RL's default DPPO objective: an
importance-weighted token policy gradient with masking of
large probability changes and a small squared log-ratio KL penalty. The value-training specific settings in Table~\ref{tab:value-hparams} are shared by
all \textsc{VF}, \pvf{}, and \tether{} runs.

\par\medskip
\noindent\begin{minipage}{\columnwidth}
  \centering
  \small
  \renewcommand{\arraystretch}{1.08}
  \refstepcounter{table}
  \label{tab:policy-hparams}
  {\small\textbf{Table~\thetable:} Policy and rollout hyperparameters shared across baseline comparisons.\par}
  \vspace{0.4em}
  \begin{tabular}{@{}p{0.31\columnwidth}p{0.62\columnwidth}@{}}
    \toprule
    Hyperparameter & Setting \\
    \midrule
    Policy learning rate & $1\times10^{-6}$ \\
    Optimizer & AdamW \\
    AdamW betas & $(0.9,\,0.999)$ \\
    Weight decay & $0.01$ \\
    Gradient clipping & Global norm $1.0$ \\
    Learning-rate schedule & Constant \\
    Sampling temperature & $1.0$ \\
    Maximum off-policyness & 8 policy updates \\
    Policy loss & PRIME-RL default DPPO+KL loss \\
    Mask and KL settings & Probability-change thresholds $0.2$ in both directions; squared log-ratio coefficient $10^{-3}$ \\
    \bottomrule
  \end{tabular}
\end{minipage}
\par\medskip

\par\medskip
\noindent\begin{minipage}{\columnwidth}
  \centering
  \small
  \renewcommand{\arraystretch}{1.08}
  \refstepcounter{table}
  \label{tab:value-hparams}
  {\small\textbf{Table~\thetable:} Value-function hyperparameters shared across value-backed runs.\par}
  \vspace{0.4em}
  \begin{tabular}{@{}p{0.31\columnwidth}p{0.62\columnwidth}@{}}
    \toprule
    Hyperparameter & Setting \\
    \midrule
    Value warmup & 20 updates before policy training \\
    Value batch size & Equal to the policy rollout batch size (differs across tasks) \\
    Replay buffer & FIFO 256 trajectories \\
    Maximum replay reuse & $N=2$ selections per trajectory \\
    Value learning rate & $1\times10^{-5}$ \\
    Value optimizer & AdamW, betas $(0.9,\,0.999)$, weight decay $0.01$ \\
    Value lr schedule & Constant \\
    Value loss & Binary cross-entropy \\
    GAE $\lambda$ & 1.0 \\
    \bottomrule
  \end{tabular}
\end{minipage}
\par\medskip

\end{document}